\documentclass[10pt,twocolumn,letterpaper]{article}

\usepackage[pagenumbers]{cvpr} 
\usepackage{multirow}
\usepackage{bbm}
\usepackage[dvipsnames]{xcolor}

\usepackage{amsmath,amsfonts,bm}

\def\eqref#1{equation~\ref{#1}}

\def\1{\bm{1}}

\DeclareMathAlphabet{\mathsfit}{\encodingdefault}{\sfdefault}{m}{sl}
\SetMathAlphabet{\mathsfit}{bold}{\encodingdefault}{\sfdefault}{bx}{n}

\def\gB{{\mathcal{B}}}

\def\gD{{\mathcal{D}}}
\def\gE{{\mathcal{E}}}

\def\gG{{\mathcal{G}}}

\def\gI{{\mathcal{I}}}

\def\gL{{\mathcal{L}}}
\def\gM{{\mathcal{M}}}
\def\gN{{\mathcal{N}}}
\def\gO{{\mathcal{O}}}
\def\gP{{\mathcal{P}}}

\def\gR{{\mathcal{R}}}

\def\gT{{\mathcal{T}}}

\def\sR{{\mathbb{R}}}

\definecolor{cvprblue}{rgb}{0.21,0.49,0.74}
\usepackage[pagebackref,breaklinks,colorlinks,citecolor=cvprblue]{hyperref}

\def\paperID{xxxx} 
\def\confName{CVPR}
\def\confYear{2024}

\def\name{GaussianEditor\xspace}

\title{GaussianEditor: Editing 3D Gaussians Delicately with Text Instructions}

\author{Junjie Wang\footnotemark[1], \;\; Jiemin Fang\footnotemark[1]\ \footnotemark[2], \;\; Xiaopeng Zhang, \;\; Lingxi Xie, \;\; Qi Tian\\
Huawei Inc.\\
\texttt{\small\{is.wangjunjie, jaminfong, zxphistory, 198808xc\}@gmail.com}  \;\;
\texttt{\small tian.qi1@huawei.com}\\
}

\begin{document}

\twocolumn[{%
\renewcommand\twocolumn[1][]{#1}%
\maketitle
\vspace{-30pt}
\begin{center}
\centering
\includegraphics[width=0.8\linewidth]{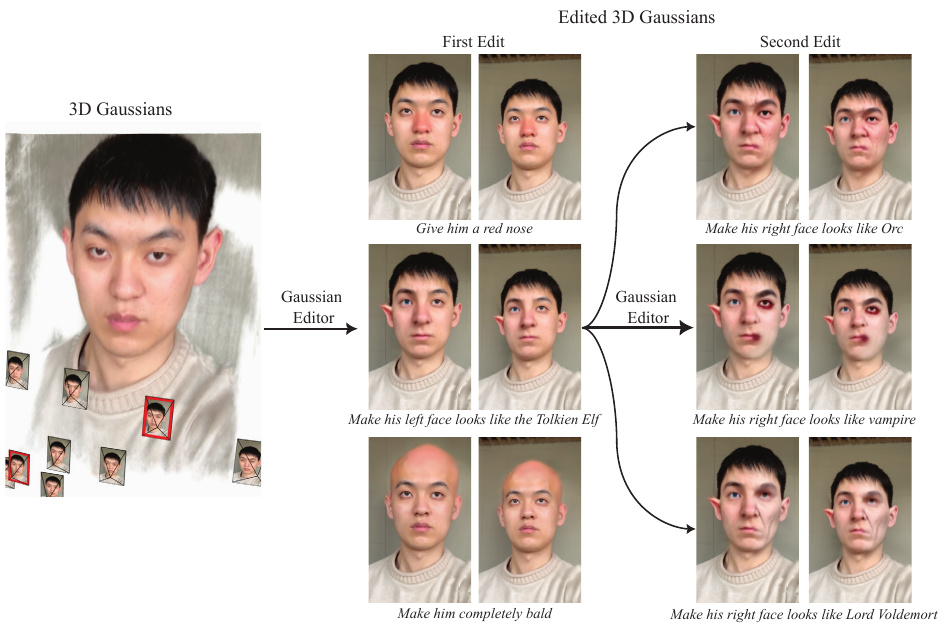}
\vspace{-5pt}
\captionof{figure}{We propose \name, an interactive framework to achieve delicate 3D scene editing following text instructions. As shown in this figure, our method can precisely control the editing region and achieve multi-round editing.
}
\label{fig: teaser}
\end{center}%
}]

{
\renewcommand{\thefootnote}{\fnsymbol{footnote}}
\footnotetext[1]{Equal contributions.}
\footnotetext[2]{Corresponding author.}
}

\begin{abstract}
\vspace{-10pt}
Recently, impressive results have been achieved in 3D scene editing with text instructions based on a 2D diffusion model. 
However, current diffusion models primarily generate images by predicting noise in the latent space, and the editing is usually applied to the whole image, which makes it challenging to perform delicate, especially localized, editing for 3D scenes. Inspired by recent 3D Gaussian splatting, we propose a systematic framework, named \name, to edit 3D scenes delicately via 3D Gaussians with text instructions. Benefiting from the explicit property of 3D Gaussians, we design a series of techniques to achieve delicate editing. Specifically, we first extract the region of interest (RoI) corresponding to the text instruction, aligning it to 3D Gaussians. The Gaussian RoI is further used to control the editing process. Our framework can achieve more delicate and precise editing of 3D scenes than previous methods while enjoying much faster training speed, \ie within 20 minutes on a single V100 GPU, more than twice as fast as Instruct-NeRF2NeRF (45 minutes -- 2 hours)\footnote{The editing time varies in different scenes according to the scene structure complexity.}. The project page is at \url{https://GaussianEditor.github.io}.
\end{abstract}

\section{Introduction}
\label{sec:intro}
Creating 3D assets has played a critical role in many applications and industries, \eg movie/game production, artistic creation, AR, VR \etc. However, this process is usually expensive and cumbersome, especially for traditional pipelines. Designers need to take a lot of labor and time to finish each step, \eg sketching, building structures, creating textures \etc. One cheap and effective way of creating high-quality 3D assets is to start from an existing scene, capturing, modeling, and editing the scene and obtaining the wanted one. This approach can be also used for user-interactive entertainment applications. 

Neural radiance field methods~\cite{mildenhall2021nerf, sun2022direct, muller2022instant, barron2021mip, barron2022mip, chen2022tensorf, yu_and_fridovichkeil2021plenoxels} have shown great power in representing 3D scenes and synthesizing novel-view images. Past years have witnessed the rapid development of NeRF and its variants, from both quality and efficiency perspectives. Editing a pre-trained NeRF model becomes a promising way to edit 3D scenes. Represented by Instruct-NeRF2NeRF~\cite{instructnerf2023}, researchers propose to use the image-conditioned 2D diffusion model, \eg InstructPix2Pix~\cite{brooks2022instructpix2pix}, to edit 3D scenes simply with text instructions. Notable results have been achieved as real scenes can be changed following the text instruction. However, current 2D diffusion models face challenges in accurately localizing editing regions, which hinders the generation of finely edited scenes due to the change of unintended regions. Even though some works~\cite{mirzaei2023watchyoursteps} propose to constrain the editing region on edited 2D images, the editing region is not accurately localized and hard to apply to the 3D representation. Besides, NeRF-based methods~\cite{nerfart, DN2N} bear coupling effects between different spatial positions, \eg different points are queried from the same MLP field (for implicit representations) or voxel vertices (for explicit representations).

Recent 3D Gaussian Splatting~\cite{kerbl20233d} (3D-GS) has been a groundbreaking work in the radiance field, which is the first to achieve a real sense of real-time rendering while enjoying high rendering quality and training speed. Besides its efficiency, we further notice its natural explicit property. 3D-GS has a great advantage for editing tasks as each 3D Gaussian exists individually. Editing 3D scenes by directly manipulating 3D Gaussians with desired constraints is easy. 

Aiming at editing 3D scenes delicately, we propose to represent the scene with 3D Gaussians, which can be edited with text instructions, and name our method as \name. \name is divided into three main parts to achieve precise control for editing regions. The first is the region of interest (RoI) extraction from the given text instruction. The instruction may be complex or indirect while this module helps extract the keywords matching the RoI for editing. The second part aligns the extracted text RoI to the 3D Gaussian space through the image space, where a grounding segmentation module is applied. The last part is to edit the original 3D Gaussians delicately with constraints in the obtained 3D Gaussian RoI. With the above processes, the region for editing can be precisely localized simply from text instructions, which constrains the 3D Gaussian updating to obtain a delicately edited new 3D scene. Besides, we enable interfaces for users to introduce more exact instructions for more delicate editing, \eg Gaussian point selecting and 3D boxes for modifying the editing regions\footnote{These additional instructions are applied to generate the man with two different edited half faces in Fig.~\ref{fig: teaser}.}.

Our contributions can be summarized as follows.
\begin{itemize}
    \item As far as we know, our \name is one of the first systematic methods to achieve delicate 3D scene editing based on 3D Gaussian splatting.
    \item A series of techniques are designed and proposed to precisely localize the editing region of interest, which are aligned and applied to 3D Gaussians. Though some sub-modules are from existing works, we believe integrating these awesome techniques to work effectively is a valuable topic, which is what we focus on in this paper.
    \item Our method achieves a series of more delicate editing results compared with the previous representative work Instruct-NeRF2NeRF~\cite{instructnerf2023} with much shorter training time (within 20 minutes \textit{v.s.} 45 minutes -- 2 hours).
\end{itemize}

\section{Related Work}
\label{sec:formatting}

\paragraph{2D Image Editing with Diffusion Models.}
Advancements in diffusion model technology~\cite{ho2020denoising, sohl2015deep}, have led to numerous generative models~\cite{saharia2022photorealistic} achieving impressive outcomes in image synthesis. Recent developments in diffusion models have demonstrated their ability to create lifelike images from arbitrary textual inputs~\cite{dhariwal2021diffusion, ho2022cascaded, saharia2022palette, saharia2022image, song2019generative}. Harnessing the robust semantic comprehension and image generation capabilities of foundational diffusion models, an escalating number of research explorations are currently employing diffusion models as a fundamental framework for implementing text-based image editing functionalities~\cite{nichol2021glide, ramesh2022hierarchical, rombach2022high, saharia2022photorealistic}. Some of these methodologies necessitate the manual provision of captions for both the original and edited images~\cite{hertz2022prompt}, while others mandate specific scenario-based training for optimization~\cite{ruiz2023dreambooth}. These requisites have rendered it arduous for ordinary users to avail themselves of such techniques. Expanding upon this foundation, iP2P~\cite{brooks2022instructpix2pix} introduces instruction-based capabilities to image editing, enabling users to simply input an image and apprise the model of the desired alterations. This user-friendly approach facilitates the democratization of image editing in a more accessible manner.

\begin{figure*}[thbp]
  \centering
  \includegraphics[width=\textwidth]{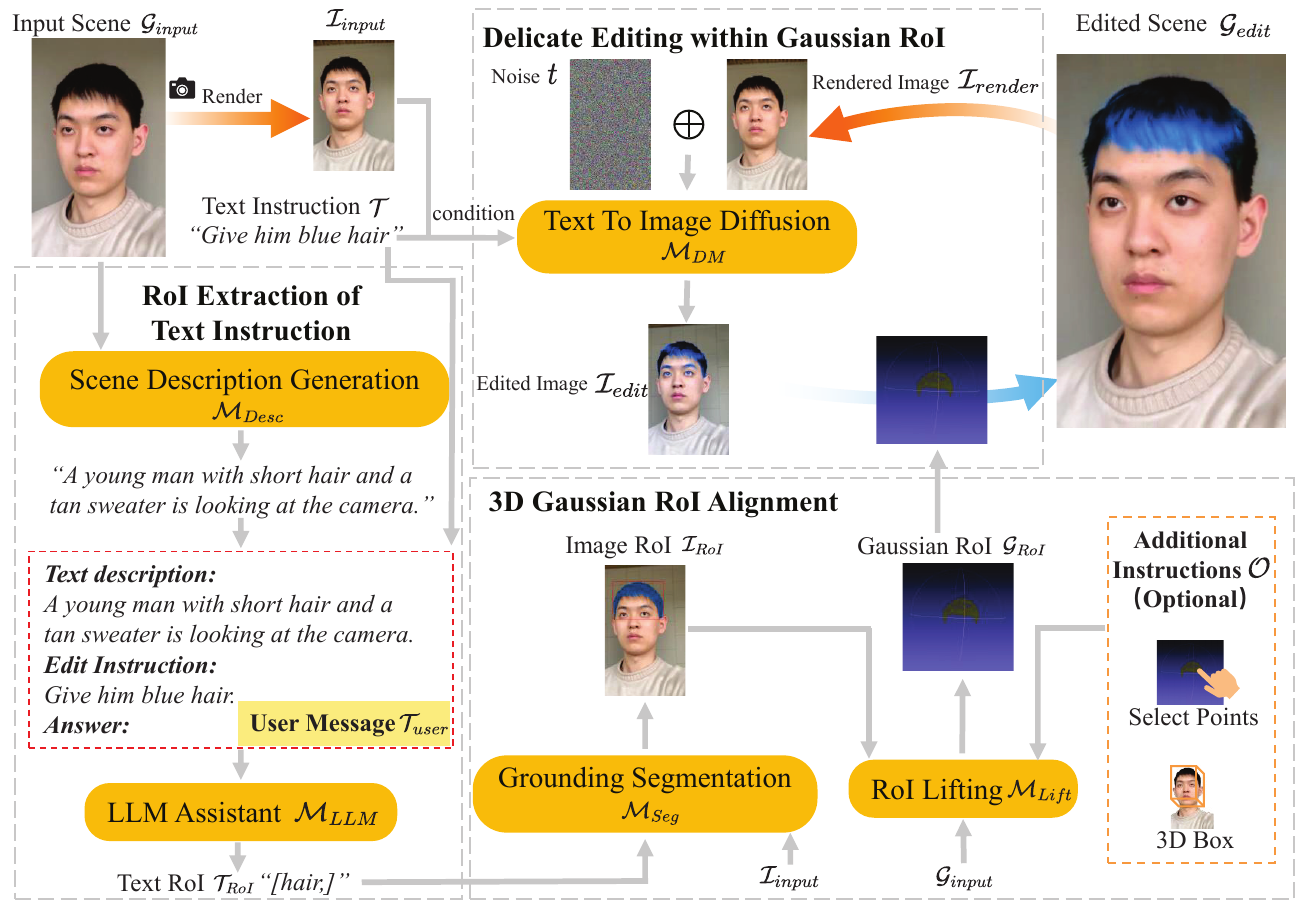}
  \vspace{-20pt}
  \caption{Our framework, named \name, consists of three key steps. First, a module $\gM_{Desc}$ is used to get the description of the input scene, which is put to an LLM assistant $\gM_{LLM}$ with the text instruction $\gT$ provided by the user to obtain the text RoI $\gT_{RoI}$. Second, a grounding segmentation module $\gM_{Seg}$ is used to convert $\gT_{RoI}$ to image RoI $\gI_{RoI}$, which is then lifted to 3D Gaussians RoI $\gG_{RoI}$ by RoI lifting $\gM_{Lift}$, where additional user instructions $\gO$ can be incorporated. Third, following the user instruction $\gT$, rendered image $\gI_{render}$ from randomly chosen views is edited by a diffusion model $\gM_{DM}$. The loss between $\gI_{render}$ and edited one $\gI_{edit}$ is calculated. Finally, gradient backpropagation and optimization are performed within the Gaussian RoI $\gG_{RoI}$ to get the edited scene  $\gG_{edit}$.}
  \label{fig: framework}
  \vspace{-15pt}
\end{figure*}

\vspace{-10pt}
\paragraph{3D Scene Editing of Radiance Fields.}
3D Scene Editing of Radiance Fields has become a popular research direction~\cite{liu2021editing, michel2022text2mesh, hong2022avatarclip, wang2022clip, kobayashi2022decomposing, tschernezki2022neural, bao2023sine, gao2023textdeformer, noguchi2021neural, liu2022nerf, li2022climatenerf, xu2022deforming, yang2022neumesh, xu2023instructp2p, li2023focaldreamer}. These methods aim to manipulate the geometry and appearance of 3D scene representations. However, editing such scenes poses challenges due to the implicit nature of traditional NeRF representations, which lack precise localization capabilities. As a result, previous works have primarily focused on achieving global style transformations of 3D scenes~\cite{nerfart, chiang2022stylizing, huang2021learning, huang2022stylizednerf, nguyen2022snerf, zhang2022arf, wu2022palettenerf}. While some efforts have been made towards object-centric scene editing\cite{zhuang2023dreameditor}, keeping the background unchanged has been a persistent challenge. For example, the recently proposed Instruct-NeRF2NeRF~\cite{instructnerf2023} implements text instruction-controlled 3D scene editing, achieving excellent editing effects while maintaining user-friendliness. However, it relies on the editing effect of 2D images, which may cause global changes to the 3D scene. A subsequent work~\cite{mirzaei2023watchyoursteps} attempts to compute the relevance map between edited and unedited images to localize the editing area. The relevance map may be unreliable when the 2D IP2P~\cite{brooks2022instructpix2pix} model fails.
Other efforts~\cite{li2023focaldreamer} rely on the user-entered 3D coordinates to determine the editing area.
The introduction of 3D Gaussians~\cite{kerbl20233d} has provided an opportunity to address this limitation. Its explicit 3D representation enables accurate selection and manipulation of editing areas. By incorporating LLMs, the whole process can be more automated.

\section{Method}
\label{sec:Method}

In this section, we first review 3D representation methods in Sec.~\ref{subsec: pre}. Subsequently, in Sec.~\ref{subsec: framework}, we overview our proposed approach, which mainly includes three modules. 
Sec.~\ref{subsec: RoI extraction} delves into the precise Region of Interest (RoI) extraction of text instructions, using scene description generation module $\gM_{Desc}$ and LLM assistant $\gM_{LLM}$. Sec.~\ref{subsec: 3D Editing Mask} introduces how to align the instruction RoI with 3D Gaussians, using grounding segmentation module $\gM_{Seg}$ and RoI lifting module $\gM_{Lift}$. Finally, Sec.~\ref{subsec: Delicate Editing} describes the delicate editing process within the obtained Gaussian RoI, using text to image diffusion model $\gM_{DM}$.
\subsection{Preliminaries} \label{subsec: pre}
\paragraph{3D Gaussian Splatting.}
3D Gaussian splatting~\cite{kerbl20233d} is a recent powerful 3D representation method. It represents the 3D scene with point-like 3D Gaussians $\gG=\{g_1,g_2...g_N\}$, where $g_i=\{\mu, \Sigma, c, \alpha\}$ and $i \in \{1, \dots, N\}$. Among them, $\mu\in \sR^3$ is the position where the Gaussian centers, $\Sigma\in\sR^7$ denotes the 3D covariance matrix, $c\in\sR^3$ is the RGB color and $\alpha\in\sR^1$ is the opacity. Benefitting from the compact representation of Gaussians and efficient differentiable rendering approach, 3D Gaussian splatting achieves real-time rendering with high quality. The splatting rendering process can be formulated as
\begin{equation}
    C=\sum_{i\in\gN}c_i\sigma_i\prod_{j=1}^{i-1}(1-\sigma_j),
    \label{eq: render}
\end{equation}
where $\sigma_i=\alpha_i e^{-\frac{1}{2}(x_i)^T\Sigma^{-1}(x_i)}$  represents the influence of the Gaussian to the image pixel and $x_i$ is the distance between the 3D point and the center of the i-th Gaussian.

\subsection{Overall Framework} \label{subsec: framework}
Given a group of 3D Gaussians $\gG_{input}$ for an input scene and a text instruction $\gT$ for editing, our Gaussian editor $\gE$ can edit the 3D Gaussians delicately into a new one, denoted as $\gG_{edit}$, with the guidance of the instruction. The whole process can be formulated as
\begin{equation}
    \gG_{edit}=\gE(\gG_{input}, \gT).
\end{equation}

Fig.~\ref{fig: framework} illustrates the overall framework of our approach, which consists of three main steps. First, the Region of Interest (RoI) is extracted from the text instruction. In this step, we employ a module named scene description generation $\gM_{Dexcription}$ to get the description of the input scene. We then input the scene description $\gT_{scene}$ and text instruction $\gT$ into a large language model assistant $\gM_{LLM}$ to determine where we should make edits in the scene. The output of this step is referred to as the instruction RoI $\gT_{RoI}$. 

The next step is the 3D Gaussian RoI alignment. We use a grounding segmentation module $\gM_{Seg}$ to convert the RoI from text space, \ie $\gT_{RoI}$, to the image space, \ie $\gI_{RoI}$. Then the image RoI $\gI_{RoI}$ is lifted to the RoI of 3D Gaussians $\gG_{RoI}$ through RoI lifting module $\gM_{Lift}$. The Gaussian RoI allows us to control the regions where edits will be applied precisely. 

The last step is delicate editing within the Gaussian RoI. In this step, we randomly sample the view to obtain the rendered image $\gI_{render}$. A 2D diffusion model $\gM_{DM}$ is used to perform the editing process on the rendered image $\gI_{render}$, with the user instruction $\gT$ and the image $\gI_{input}$ of input scene as conditions. The resulting edited image is denoted as $\gI_{edit}$. Subsequently, we calculate the loss between $\gI_{edit}$ and $\gI_{render}$ and make gradient back-propagation within $\gG_{RoI}$. This implies that only the regions specified by the RoI can receive corresponding gradients during the back-propagation process. Finally, optimization is executed based on these gradients. The final optimized scene representation $\gG_{edit}$ is obtained through several rounds of iterative optimization.

\begin{figure}[t]
   \centering
   \includegraphics[width= 0.95 \linewidth]{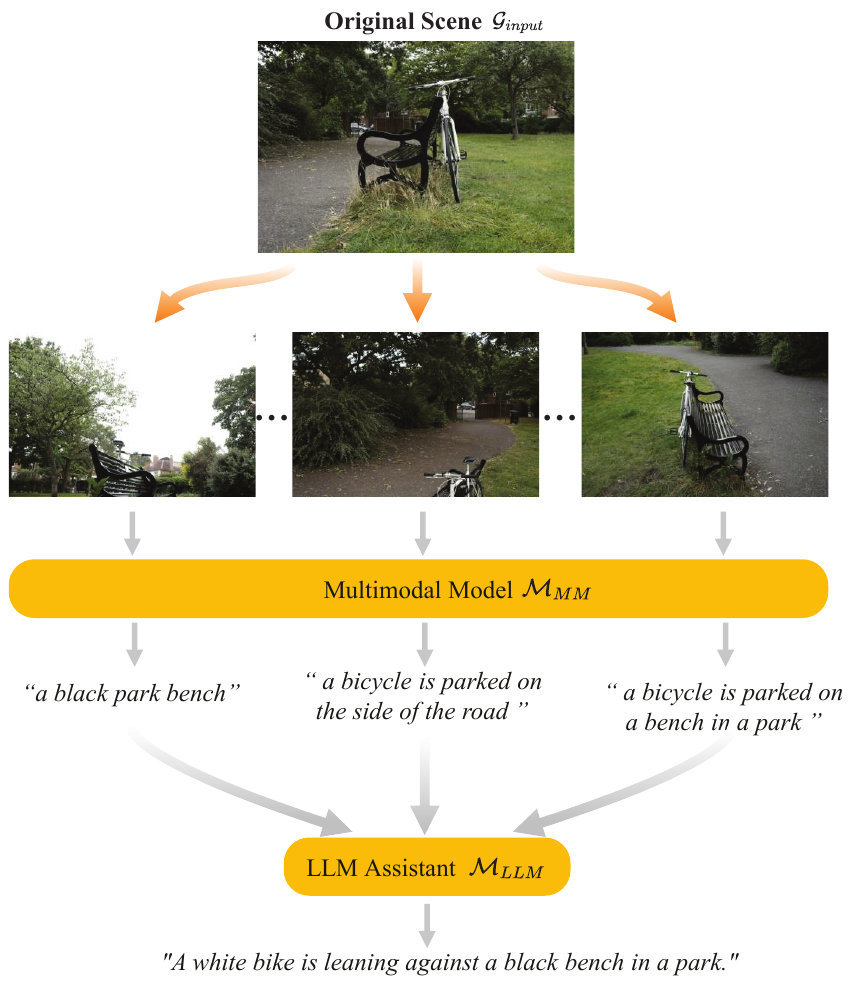}
    \vspace{-5pt}
   \caption{The process of obtaining scene description.}
   \label{fig: scene dsc}
    \vspace{-12pt}
\end{figure}

\subsection{RoI Extraction of Text Instruction} \label{subsec: RoI extraction}
The instruction RoI is extracted for the editing regions from both the input 3D scene $\gG_{input}$ and the text instruction $\gT$ provided by the user. To achieve this, we employ a multimodal model $\gM_{MM}$ in conjunction with the large language model assistant $\gM_{LLM}$. The first step is the scene description generation $\gM_{Desc}$, which aims to get the scene description $\gT_{scene}$ from 3D Gaussians $\gG_{input}$:
\begin{equation}
    \gT_{scene}=\gM_{Desc}(\gG_{input}).
\end{equation}
The process of the scene description generation $\gM_{Desc}$ is shown in Fig.~\ref{fig: scene dsc}.
By leveraging the technique of differentiable splatting as shown in Eq.~\ref{eq: render}, a set of 2D image samples $\{\gI_{sample}\}$ are generated and then inputted into a multimodal model $\gM_{MM}$ to generate corresponding text descriptions $\{\gT_{sample}\}$:
\begin{equation}
    \gT_{sample}=\gM_{MM}(\gP_{MM}, \gI_{sample} ),
\end{equation}
where $\gP_{MM}$ is a prompt, such as ``\texttt{What is the content of the image}", for multimodal model $\gM_{MM}$ to get precise description. Subsequently, these descriptions $\{\gT_{sample}\}$ are fed into a large language model $\gM_{LLM}$, which is specifically instructed by a prompt $\gP_{merge}$ to merge descriptions of diverse views into one detailed scene description $\gT_{scene}$:
\begin{equation}
    \gT_{scene}=\gM_{LLM}(\gP_{merge}, \{\gT_{sample}\}).
\end{equation}

After that, the scene description $\gT_{scene}$ and the user instruction $\gT$ are combined with a predefined template $\gT_{template}$: ``\texttt{Text description:} 
$\gT_{scene}$ \texttt{Edit Instruction:} $\gT$
\texttt{Answer:}" to form the user message $\gT_{user}$=$\gT_{template}(\gT_{scene}, \gT)$. The LLM model $\gM_{LLM}$ is used to extract the instruction RoI $\gT_{RoI}$ from user message $\gT_{user}$ with a new prompt $\gP_{extract}$:
\begin{equation}
    \gT_{RoI}=\gM_{LLM}(\gP_{extract}, \gT_{user}).
\end{equation}

\subsection{3D Gaussian RoI Alignment} \label{subsec: 3D Editing Mask}

To confine the 3D editing region within the instruction RoI, 3D Gaussian RoI $\gG_{RoI}$ is aligned with the text RoI $\gT_{RoI}$. First, The RoI in the text space is transformed into the image space via a grounding segmentation module $\gM_{Seg}$:
\begin{equation}
    \gI_{RoI}=\gM_{Seg}(\gI_{input}, \gT_{RoI}),
\end{equation}
where $\gI_{input}$ is rendered image of the input scene $\gG_{input}$.

Then we lift the the RoI $\gI_{RoI}$ in the image space to 3D Gaussian $\gG_{RoI}$ through training. To achieve this, an additional RoI attribute $r\in\sR^1$ was added to 3D Gaussian $g_i=\{\mu_i, \Sigma_i, c_i, \alpha_i, r_i\}$. $r$ is initialized to 0, which means it is not in the Gaussians RoI, and 1 means it is inside the RoI. The set of $r$ is denoted as $\gR\in\sR^{\gN, 1}$, where the $\gN$ is the number of 3D Gaussians $\gG_{input}$.

Then the color $c_i$ in Eq.~\ref{eq: render} was rewritten with $r_i$ to get the rendered RoI $\gI_{RoI}^{render}$:
\begin{equation}
    \gI_{RoI}^{render}=\sum_{i\in\gN}r_i\sigma_i\prod_{j=1}^{i-1}(1-\sigma_j).
    \label{eq: render RoI}
\end{equation}

Taking inspiration from SA3D~\cite{cen2023segment}, to get the trained Gaussians RoI $\gG_{RoI}^{train}$, we adopt a similar loss function to supervise the training process:
\begin{equation}
  \mathcal{L}_{proj} = \lambda_{1}\sum(\gI_{RoI}^{render} \cdot \gI_{RoI}) + \lambda_{2} \sum((1 - \gI_{RoI}) \cdot \gI_{RoI}^{render}),
  \label{eq:lproj}
\end{equation}
where $\lambda_{1}$ and $\lambda_{2}$ are hyperparameters. The $r$ in Eq.~\ref{eq: render RoI} is updated via $r\gets r-\eta\frac{\partial \mathcal{L}_{proj}}{\partial r}$ with gradient descent, where $\eta$ denotes the learning rate. Eq.~\ref{eq:lproj} encourages rendered RoI to cover the Image RoI and not exceed it. Additionally, the user can modify the trained Gaussian RoI $\gG_{RoI}^{train}$ by giving added Gaussian RoI $\gG_{RoI}^{add}$, deleted Gaussian RoI $\gG_{RoI}^{del}$ and 3D box $\gB_{3D}$:
\begin{equation}
    \gG_{RoI}=(\gG_{RoI}^{train}\cup~\gG_{RoI}^{add}-\gG_{RoI}^{del})\cap~\gB_{3D},
\end{equation}
$\gG_{RoI}^{add}$ represents the 3D Gaussians user wants to edit, $\gG_{RoI}^{del}$ means 3D Gaussians user wants to keep from editing, $\gB_{3D}$ is the coordinates of 3D cuboid it limits RoI to inside the box. $\gG_{RoI}$ is the aligned RoI with the text RoI. For example, when editing the left face of the man in Fig.~\ref{fig: teaser}, grounding segmentation failed to ground ``left face'', instead, it grounded the whole face. In this scenario, the user can use the interactive interface to set the right face as $\gG_{RoI}^{del}$ or enter the rectangular box where the left face is located as $\gB_{3D}$. The lifting process $\gM_{Lift}$ can be represented as:
\begin{equation}
    \gG_{RoI}=\gM_{Lift}(\gI_{RoI}, \gO),
\end{equation}
where $\gO=\{ \gG_{RoI}^{add}, \gG_{RoI}^{del}, \gB_{3D} \}$ is optional instructions.

\subsection{Delicate Editing within Gaussian RoI} \label{subsec: Delicate Editing}
To achieve delicate editing in 3D scenes, we use the Gaussian RoI to constrain the editing area. In particular, we randomly sample viewpoints from the 3D scene and render 2D image $\gI_{render}$. After that, $\gI_{render}$ and noise level $t$ are put into 2D diffusion model $\gM_{DM}$, with the user instruction $\gT$ and image $\gI_{input}$ of input scene as conditions, to get edited image $\gI_{edit}$:
\begin{equation}
    \gI_{edit}=\gD(\gI_{render}, t;\gT,\gI_{input}),
\end{equation}
where $t$ is a randomly chosen noise level from $[t_{min}, t_{max}]$.

Similar to 3D-GS~\cite{kerbl20233d}, we apply the $\mathcal{L}_\mathbbm{1}$ and D-SSIM loss functions during editing.
\begin{equation}
    \mathcal{L}=(1-\beta)\mathcal{L}_\mathbbm{1}+\beta \mathcal{L}_{D-SSIM}.
    \label{eq: edit loss}
\end{equation}
the two losses are calculated between the 2D edited image $\gI_{ed}$ and the rendered image $\gI_{rd}$. Then, gradient backpropagation is performed within Gaussian RoI $\gG_{RoI}$:
\begin{equation}
\nabla\gG=\frac{\partial \gL}{\partial \gG} \cdot \gR,
\label{eq: RoI control gradient}
\end{equation}
\begin{figure*}[thbp]
  \centering
  \includegraphics[width=0.98\textwidth]{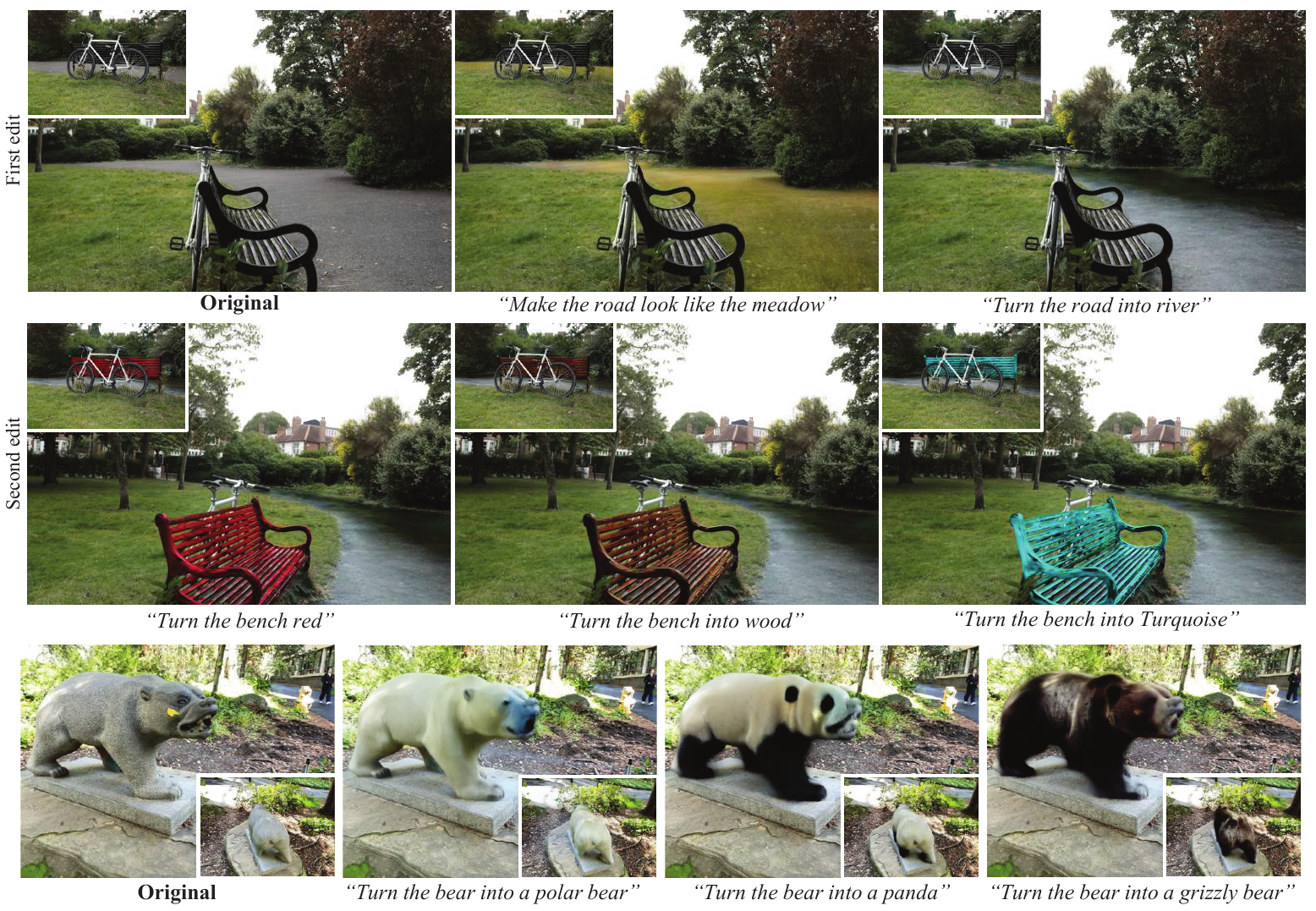}
  \vspace{-10pt}
  \caption{Qualitative results on outdoor scenes. Our method supports separate foreground and background editing in real-world scenes.}
  \label{fig: bicycle_and_bear}
  \vspace{-10pt}
\end{figure*}
where $\gR$ is the set of RoI attributes. That means only Gaussians in RoI can receive gradients. Finally, we utilize the Adam algorithm to optimize the 3D Gaussians. After many rounds of training, the edited 3D scene $\gG_{edit}$ is obtained.

\section{Experiments}
\label{sec:Experiments}

\subsection{Implementation Details}
Our method is implemented in PyTorch~\cite{paszke2019pytorch} and CUDA, based on 3D Gaussian splatting. The multimodal model we used in our method is BLIP2~\cite{li2023blip}, and we use GPT-3.5 Turbo to ground the text ROI. For grounding segmentation, We use the cascade strategy, \ie first using Grounding DINO~\cite{liu2023grounding} to get the box on the image corresponding to the text, and then using SAM~\cite{kirillov2023segment} to get the corresponding image RoI. The 2D diffusion model used in our method is Instruct Pix2Pix~\cite{brooks2022instructpix2pix}. We leave more details in the Appendix.

\subsection{Qualitative Evaluation}
\begin{figure*}[thbp]
  \centering
  \includegraphics[width=\textwidth]{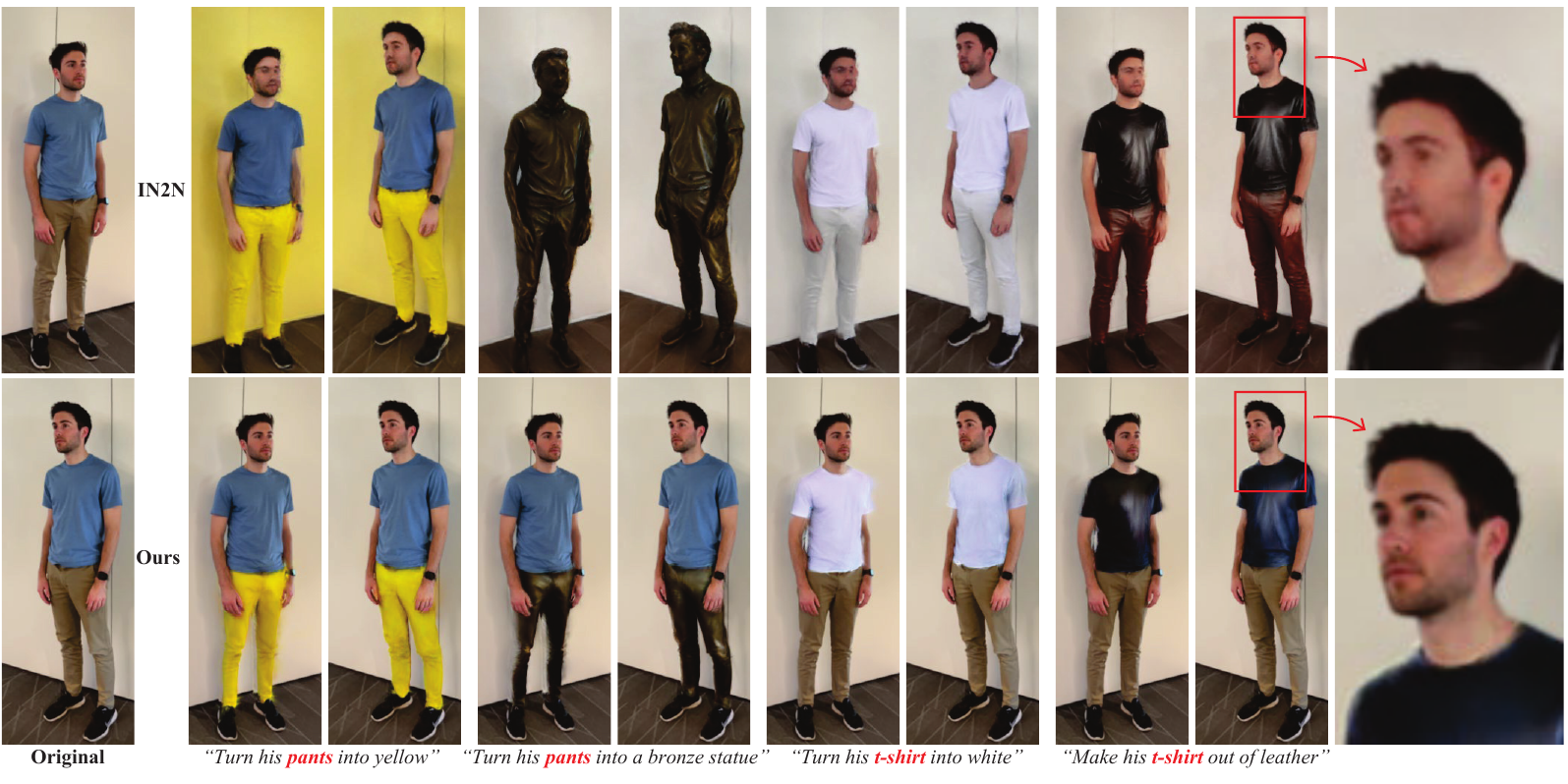}
  \vspace{-23pt}
  \caption{Comparisons with Instruct-NeRF2NeRF (IN2N)~\cite{instructnerf2023} on the scene presented in their paper.}
  \vspace{-8pt}
  \label{fig: compare with in2n}
\end{figure*}

\paragraph{Visualization Results.}
In Fig.~\ref{fig: teaser} and Fig.~\ref{fig: bicycle_and_bear}, we present the visual results of \name, demonstrating the precise editing effects while ensuring 3D consistency. Fig.~\ref{fig: teaser} shows the editing capabilities for characters. The first column displays the original scenes. In the second column, the first row ``\texttt{Give him a red nose}" illustrates color-changing ability, while the third row, ``\texttt{Make him completely bald}", showcases capabilities of retexturing and slight geometry editing. The second row in the second column demonstrates precise editing ability by exclusively editing the left side of the face. Based on that, we achieve editing in the third column, focusing on the right side of the face, showcasing the ability of multi-round edits, and accurately fulfilling user instructions. Fig.~\ref{fig: bicycle_and_bear} showcases the precise editing capabilities in open 3D scenes. In the upper portion, the bicycle scene allows us to accurately locate the position of the road and edit its texture, transforming it into the grass, a river. In the experiment where we change the texture to a river, our method accurately constructs the reflection, making it appear realistic. Based on editing the road into a river, we further edited the bench, proving that our method can achieve multiple rounds of editing. The lower portion demonstrates the results of editing the bear, which fully preserves the original appearance of the background area and focuses the edits on the bear.
\vspace{-15pt}
\paragraph{Comparisons with Instruct-NeRF2NeRF.} Fig.~\ref{fig: compare with in2n} compares the results of our method with those of IN2N~\cite{instructnerf2023}, on the scenes presented in IN2N. From the figure, it is evident that our method changes the texture of the pants without affecting the clothes, and vice versa, demonstrating the effectiveness of our method in distinguishing different objects within the foreground. Additionally, when editing the clothes and pants, the background remains unaffected, indicating our method's effective separation of foreground and background. Furthermore, the last column reveals that IN2N, limited by 2D diffusion, distorts the face, while our method maintains a superior rendering quality of faces.

\begin{figure*}[thbp]
  \centering
  \includegraphics[width=\textwidth]{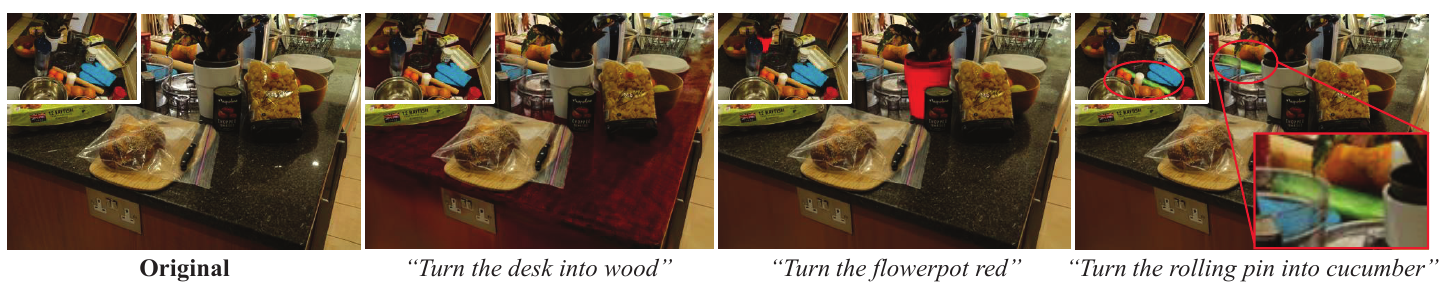}
  \vspace{-20pt}
  \caption{Qualitative results on complex multi-object scenes. The background ``\texttt{desk}", the foreground ``\texttt{flower pot}", and the multi-view blocked foreground ``\texttt{rolling pin}" are edited separately.}
  \label{fig: counter}
  \vspace{-15pt}
\end{figure*}
\vspace{-10pt}

\paragraph{Complex Multi-Object Scenes.} Furthermore, we present the results of our editing in a complex scene featuring multiple objects, as depicted in Fig~\ref{fig: counter}. Three distinct object types are selected for editing purposes. The first type is the background, which is the desktop in this scene. We successfully transformed the desktop into a wooden material using a caption-based approach. The edited result exhibits a distinct wood texture. The second object type is a foreground object, the flowerpot. We opted to change the color of the flowerpot to red, and the outcome was highly successful. Lastly, the most intricate editing task involved the rolling pin, which was occluded by multiple objects from various perspectives. As shown in the lower right corner of the picture, we managed to edit it into a cucumber without impacting the other objects.

\begin{table}[htbp]
\centering
\caption{Quantitative evaluation on the bicycle scene of the Mip-NeRF360 dataset~\cite{barron2022mip}.}
\vspace{-5pt}
\begin{tabular}{l|c|c|c|c}
\toprule
\textbf{Methods} & \textbf{CTIDS}$\uparrow$
 &  \textbf{IIS}$\uparrow$ & \textbf{FID}$\downarrow$ & \textbf{Time}$\downarrow$ \\
\midrule
IN2N~\cite{instructnerf2023} & 0.22 & 0.85 & 103 & 51 min\\
Ours-DVGO~\cite{sun2022direct} & 0.11 & 0.82 & 148 & 40 min\\
Ours-3DGS & \textbf{0.28} & \textbf{0.95} & \textbf{51} & \textbf{20} min\\
\bottomrule
\end{tabular}
\vspace{-10pt}
\label{tab: quantitative}
\end{table}

\subsection{Quantitative Evaluation} \label{subsec: quantitative}
\paragraph{Metric Comparisons.} Table~\ref{tab: quantitative} shows quantitative results on the bicycle scene of the Mip-NeRF360 dataset~\cite{barron2022mip}, comparing with IN2N~\cite{instructnerf2023} and Direct Voxel Grid Optimization (DVGO)~\cite{sun2022direct} as the representation. The metrics include CLIP Text-Image Direction Similarity (CTIDS), Image-Image Similarity (IIS), FID, and training time. \name achieves the best results in all metrics. The test data is shown in the supplementary material.
\vspace{-10pt}
\paragraph{User Study.} We perform a user study comparing with IN2N on the bear scene in Fig.~\ref{fig: bicycle_and_bear} and the human scene in Fig.~\ref{fig: compare with in2n}, involving 21 participants. \name gets a 87.07\% voting percentage, while IN2N gets 12.93\%.

\subsection{Ablation Study and Analysis}
\begin{figure}[t]
   \centering
   \includegraphics[width= 0.95 \linewidth]{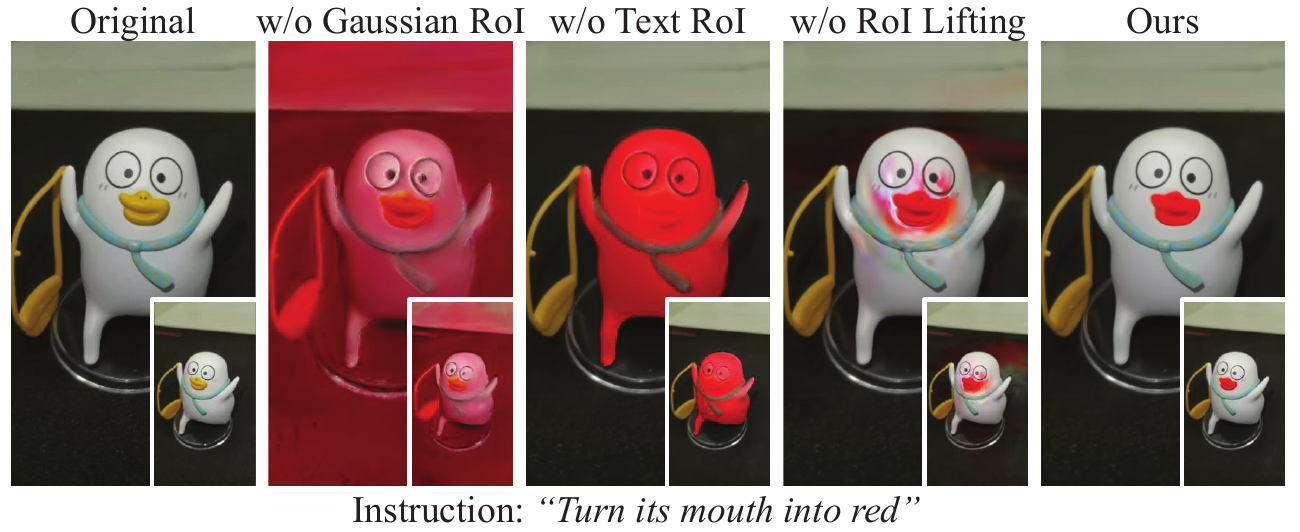}
    \vspace{-7pt}
   \caption{Ablation experiment of RoI.}
   \vspace{-20pt}
   \label{fig: ablation}
\end{figure}

\paragraph{Ablation of Gaussian RoI, Text RoI, RoI Lifting.} 
To validate the effectiveness of each module in our framework, we design three variant approaches: (1) w/o Gaussian RoI: We discontinued the use of Gaussian RoI $\gG_{RoI}$ to control the gradients of Gaussian points, as mentioned in Eq.~\ref{eq: RoI control gradient}. (2) w/o Text ROI: In this scenario, we ceased the selection of text ROI $\gT_{RoI}$ using LLM assistant $\gM_{LLM}$. Instead, all the words in the user's instruction are put to $\gM_{Seg}$ to get $\gI_{RoI}$. (3) w/o RoI lifting: Instead of lifting the image RoI $\gI_{RoI}$ to 3D Gaussians, the image RoI $\gI_{RoI}$ is used to govern the calculation of the loss. That is, only the pixels within the image RoI $\gI_{RoI}$ are taken into account for the loss computation.
Fig.~\ref{fig: ablation} showcases the outcomes of our ablation experiment, which aimed to edit the doll based on the instruction ``\texttt{Turn its mouth into red.}" The results reveal the following findings. (1) When the Gaussian RoI is not used, the 3D scene is all turned red because the 2D diffusion fails to control the editing area. (2) In cases where text ROI $\gT_{RoI}$ is not utilized, the grounding segmentation model tends to segment the entire foreground object, leading to the doll being entirely edited to red. (3) When RoI lifting $\gM_{Lift}$ is not employed, the doll's mouth is successfully turned red, but other facial areas are also affected. Because the grounding segmentation model may fail to parse specific views, noise exists on the image RoI $\gI_{RoI}$. Consequently, leakage occurs during the editing process. Our proposed RoI lifting module effectively addresses this issue during training. In conclusion, our ablation experiment demonstrates the effectiveness of several RoI-related modules in our method. 

\begin{figure}[t]
   \centering
   \includegraphics[width= 0.95 \linewidth]{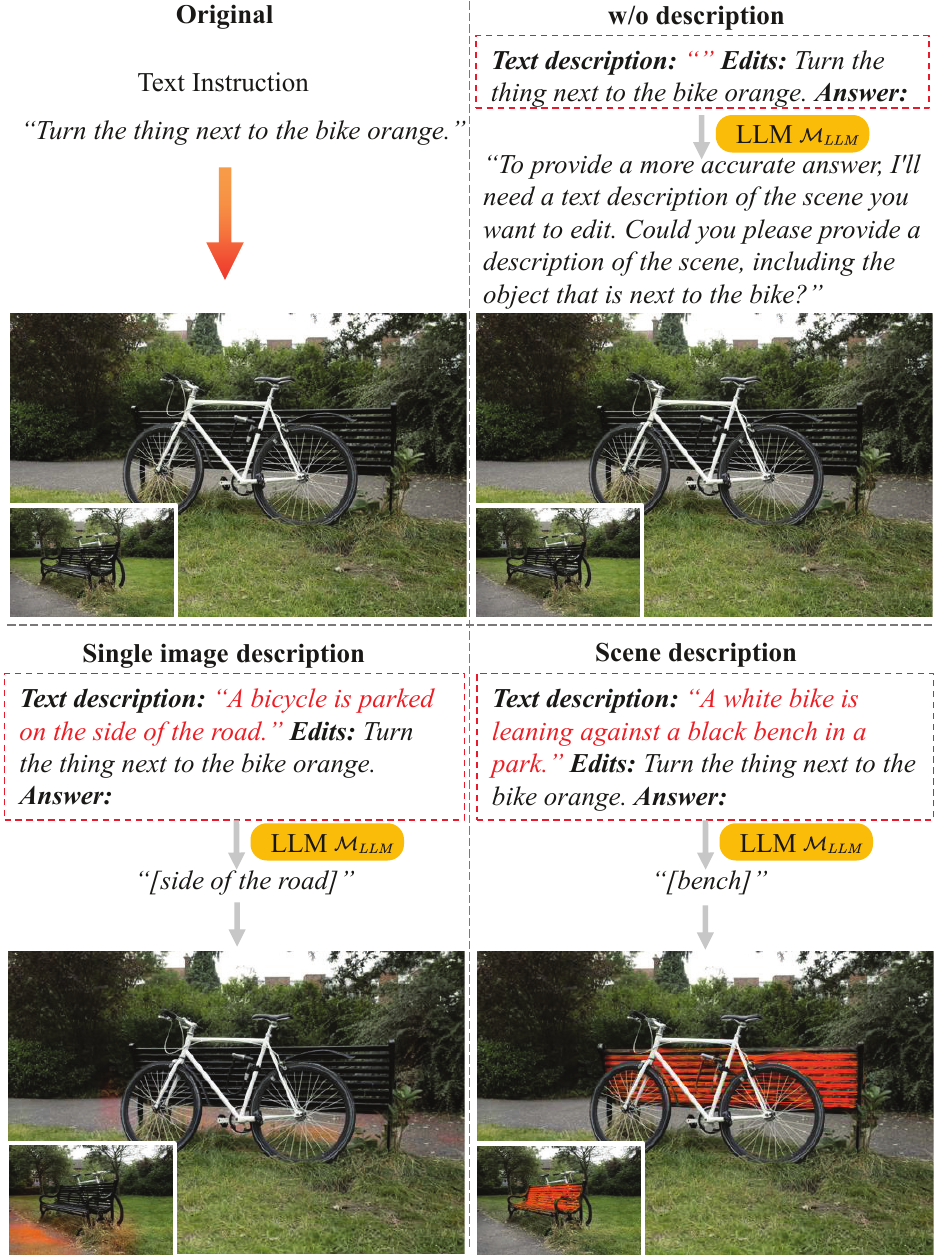}
    \vspace{-5pt}
   \caption{Ablation results about the scene description generation.}
   \label{fig: ablation scene description}
    \vspace{-18pt}
\end{figure}

\vspace{-14pt}
\paragraph{Ablation of Scene Description Generation.}
We further conduct experiments to evaluate the role of scene description generation, employing three distinct experimental setups. The first one composes the user message $\gT_{user}$, without employing scene description. The second method randomly samples a view and extracts the corresponding image's text description as the scene description. The third one represents the complete version of our approach.
The test scene involves a park where a bike and a bench are positioned closely together. The editing instruction is ``\texttt{Turn the thing next to the bike orange}". The obtained results are presented in Fig.~\ref{fig: ablation scene description}. As shown in the image, when scene description is not employed, the LLM fails to acquire the Text ROI according to the user instructions, resulting in editing failure. The second one randomly samples images to obtain scene descriptions, resulting in incomplete descriptions and leading to an incorrect text ROI prediction by the LLM. Consequently, the final editing result turns the road into an orange color. In contrast, our method flawlessly executes the editing task. This success can be attributed to scene description generation, which obtains an accurate text description encompassing the relative positional relationship between the bicycle and the bench. This enables the LLM to analyze and determine the user's intention to edit the bench. Consequently, the desired color change of the bench is successfully implemented.

\subsection{Limitations}
Although our framework has solved some problems inherited from the integrated sub-modules, \eg noise in the results of grounding segmentation, there are still some problems that the current system cannot completely avoid. In scene description generation, the descriptions from different views of the same object may differ from each other. When the differences are large enough, the LLM may misunderstand these descriptions as those from multiple objects. This issue does not affect the results in the current experiment, but we would like to optimize this in the future. In addition, our system cannot achieve good editing results in scenes where the grounding segmentation or diffusion model completely fails, such as drastic geometric editing. 

\section{Conclusion}
This paper proposes a systematic framework, named \name, for text-guided delicate 3D scene editing. As we know, \name is one of the first works to edit 3D Gaussians, taking advantage of the explicit property of 3D Gaussians and making it easy to control the editing area precisely. Several techniques are proposed to achieve delicate editing, including extracting instruction RoI from texts, aligning the RoI to 3D Gaussians, and editing the scene with the Gaussian RoI. \name achieves notably more delicate editing results than IN2N~\cite{instructnerf2023} with much shorter training time (within 20 minutes \textit{v.s.} 45 minutes -- 2 hours). Noticing recent works~\cite{luiten2023dynamic, wu20234dgaussians, yang2023deformable3dgs, yang2023gs4d} have extended Gaussian splatting to dynamic scenes, we leave the delicate editing in dynamic scenes as future work.

{
    \small
    \bibliographystyle{ieeenat_fullname}
    \bibliography{main}

\begin{thebibliography}{59}
\providecommand{\natexlab}[1]{#1}
\providecommand{\url}[1]{\texttt{#1}}
\expandafter\ifx\csname urlstyle\endcsname\relax
  \providecommand{\doi}[1]{doi: #1}\else
  \providecommand{\doi}{doi: \begingroup \urlstyle{rm}\Url}\fi

\bibitem[Bao et~al.(2023)Bao, Zhang, Yang, Fan, Yang, Bao, Zhang, and Cui]{bao2023sine}
Chong Bao, Yinda Zhang, Bangbang Yang, Tianxing Fan, Zesong Yang, Hujun Bao, Guofeng Zhang, and Zhaopeng Cui.
\newblock Sine: Semantic-driven image-based nerf editing with prior-guided editing field.
\newblock In \emph{CVPR}, 2023.

\bibitem[Barron et~al.(2021)Barron, Mildenhall, Tancik, Hedman, Martin-Brualla, and Srinivasan]{barron2021mip}
Jonathan~T Barron, Ben Mildenhall, Matthew Tancik, Peter Hedman, Ricardo Martin-Brualla, and Pratul~P Srinivasan.
\newblock Mip-nerf: A multiscale representation for anti-aliasing neural radiance fields.
\newblock In \emph{ICCV}, 2021.

\bibitem[Barron et~al.(2022)Barron, Mildenhall, Verbin, Srinivasan, and Hedman]{barron2022mip}
Jonathan~T Barron, Ben Mildenhall, Dor Verbin, Pratul~P Srinivasan, and Peter Hedman.
\newblock Mip-nerf 360: Unbounded anti-aliased neural radiance fields.
\newblock In \emph{CVPR}, 2022.

\bibitem[Brooks et~al.(2023)Brooks, Holynski, and Efros]{brooks2022instructpix2pix}
Tim Brooks, Aleksander Holynski, and Alexei~A Efros.
\newblock Instructpix2pix: Learning to follow image editing instructions.
\newblock In \emph{CVPR}, 2023.

\bibitem[Cen et~al.(2023)Cen, Zhou, Fang, Yang, Shen, Xie, Jiang, Zhang, and Tian]{cen2023segment}
Jiazhong Cen, Zanwei Zhou, Jiemin Fang, Chen Yang, Wei Shen, Lingxi Xie, Dongsheng Jiang, Xiaopeng Zhang, and Qi Tian.
\newblock Segment anything in 3d with nerfs.
\newblock In \emph{NeurIPS}, 2023.

\bibitem[Chen et~al.(2022)Chen, Xu, Geiger, Yu, and Su]{chen2022tensorf}
Anpei Chen, Zexiang Xu, Andreas Geiger, Jingyi Yu, and Hao Su.
\newblock Tensorf: Tensorial radiance fields.
\newblock In \emph{ECCV}, 2022.

\bibitem[Chiang et~al.(2022)Chiang, Tsai, Tseng, Lai, and Chiu]{chiang2022stylizing}
Pei-Ze Chiang, Meng-Shiun Tsai, Hung-Yu Tseng, Wei-Sheng Lai, and Wei-Chen Chiu.
\newblock Stylizing 3d scene via implicit representation and hypernetwork.
\newblock In \emph{Proceedings of the IEEE/CVF Winter Conference on Applications of Computer Vision}, 2022.

\bibitem[Dhariwal and Nichol(2021)]{dhariwal2021diffusion}
Prafulla Dhariwal and Alexander Nichol.
\newblock Diffusion models beat gans on image synthesis.
\newblock \emph{Advances in neural information processing systems}, 2021.

\bibitem[Fang et~al.(2023)Fang, Wang, Yang, Yang, Tsai, Ding, Yang, and Zhou]{DN2N}
Shuangkang Fang, Yufeng Wang, Yi Yang, Yi Yang, Yi-Hsuan Tsai, Wenrui Ding, Ming-Hsuan Yang, and Shuchang Zhou.
\newblock Text-driven editing of 3d scenes without retraining.
\newblock \emph{Arxiv preprint arXiv:2309.04917}, 2023.

\bibitem[Gao et~al.(2023)Gao, Aigerman, Groueix, Kim, and Hanocka]{gao2023textdeformer}
William Gao, Noam Aigerman, Thibault Groueix, Vova Kim, and Rana Hanocka.
\newblock Textdeformer: Geometry manipulation using text guidance.
\newblock In \emph{ACM SIGGRAPH 2023 Conference Proceedings}, 2023.

\bibitem[Haque et~al.(2023)Haque, Tancik, Efros, Holynski, and Kanazawa]{instructnerf2023}
Ayaan Haque, Matthew Tancik, Alexei Efros, Aleksander Holynski, and Angjoo Kanazawa.
\newblock Instruct-nerf2nerf: Editing 3d scenes with instructions.
\newblock In \emph{CVPR}, 2023.

\bibitem[Hertz et~al.(2022)Hertz, Mokady, Tenenbaum, Aberman, Pritch, and Cohen-Or]{hertz2022prompt}
Amir Hertz, Ron Mokady, Jay Tenenbaum, Kfir Aberman, Yael Pritch, and Daniel Cohen-Or.
\newblock Prompt-to-prompt image editing with cross attention control.
\newblock \emph{arXiv preprint arXiv:2208.01626}, 2022.

\bibitem[Ho et~al.(2020)Ho, Jain, and Abbeel]{ho2020denoising}
Jonathan Ho, Ajay Jain, and Pieter Abbeel.
\newblock Denoising diffusion probabilistic models.
\newblock \emph{NeurIPS}, 2020.

\bibitem[Ho et~al.(2022)Ho, Saharia, Chan, Fleet, Norouzi, and Salimans]{ho2022cascaded}
Jonathan Ho, Chitwan Saharia, William Chan, David~J Fleet, Mohammad Norouzi, and Tim Salimans.
\newblock Cascaded diffusion models for high fidelity image generation.
\newblock \emph{The Journal of Machine Learning Research}, 2022.

\bibitem[Hong et~al.(2022)Hong, Zhang, Pan, Cai, Yang, and Liu]{hong2022avatarclip}
Fangzhou Hong, Mingyuan Zhang, Liang Pan, Zhongang Cai, Lei Yang, and Ziwei Liu.
\newblock Avatarclip: Zero-shot text-driven generation and animation of 3d avatars.
\newblock \emph{arXiv preprint arXiv:2205.08535}, 2022.

\bibitem[Huang et~al.(2021)Huang, Tseng, Saini, Singh, and Yang]{huang2021learning}
Hsin-Ping Huang, Hung-Yu Tseng, Saurabh Saini, Maneesh Singh, and Ming-Hsuan Yang.
\newblock Learning to stylize novel views.
\newblock In \emph{Proceedings of the IEEE/CVF International Conference on Computer Vision}, 2021.

\bibitem[Huang et~al.(2022)Huang, He, Yuan, Lai, and Gao]{huang2022stylizednerf}
Yi-Hua Huang, Yue He, Yu-Jie Yuan, Yu-Kun Lai, and Lin Gao.
\newblock Stylizednerf: consistent 3d scene stylization as stylized nerf via 2d-3d mutual learning.
\newblock In \emph{CVPR}, 2022.

\bibitem[Kerbl et~al.(2023)Kerbl, Kopanas, Leimk{\"u}hler, and Drettakis]{kerbl20233d}
Bernhard Kerbl, Georgios Kopanas, Thomas Leimk{\"u}hler, and George Drettakis.
\newblock 3d gaussian splatting for real-time radiance field rendering.
\newblock \emph{ACM Transactions on Graphics (ToG)}, 2023.

\bibitem[Kirillov et~al.(2023)Kirillov, Mintun, Ravi, Mao, Rolland, Gustafson, Xiao, Whitehead, Berg, Lo, et~al.]{kirillov2023segment}
Alexander Kirillov, Eric Mintun, Nikhila Ravi, Hanzi Mao, Chloe Rolland, Laura Gustafson, Tete Xiao, Spencer Whitehead, Alexander~C Berg, Wan-Yen Lo, et~al.
\newblock Segment anything.
\newblock \emph{arXiv preprint arXiv:2304.02643}, 2023.

\bibitem[Kobayashi et~al.(2022)Kobayashi, Matsumoto, and Sitzmann]{kobayashi2022decomposing}
Sosuke Kobayashi, Eiichi Matsumoto, and Vincent Sitzmann.
\newblock Decomposing nerf for editing via feature field distillation.
\newblock \emph{Advances in Neural Information Processing Systems}, 2022.

\bibitem[Li et~al.(2023{\natexlab{a}})Li, Li, Savarese, and Hoi]{li2023blip}
Junnan Li, Dongxu Li, Silvio Savarese, and Steven Hoi.
\newblock Blip-2: Bootstrapping language-image pre-training with frozen image encoders and large language models.
\newblock \emph{arXiv preprint arXiv:2301.12597}, 2023{\natexlab{a}}.

\bibitem[Li et~al.(2022)Li, Lin, Forsyth, Huang, and Wang]{li2022climatenerf}
Yuan Li, Zhi-Hao Lin, David Forsyth, Jia-Bin Huang, and Shenlong Wang.
\newblock Climatenerf: Physically-based neural rendering for extreme climate synthesis.
\newblock \emph{arXiv e-prints}, 2022.

\bibitem[Li et~al.(2023{\natexlab{b}})Li, Dou, Shi, Lei, Chen, Zhang, Zhou, and Ni]{li2023focaldreamer}
Yuhan Li, Yishun Dou, Yue Shi, Yu Lei, Xuanhong Chen, Yi Zhang, Peng Zhou, and Bingbing Ni.
\newblock Focaldreamer: Text-driven 3d editing via focal-fusion assembly.
\newblock \emph{arXiv preprint arXiv:2308.10608}, 2023{\natexlab{b}}.

\bibitem[Liu et~al.(2022)Liu, Shen, Chen, et~al.]{liu2022nerf}
Hao-Kang Liu, I Shen, Bing-Yu Chen, et~al.
\newblock Nerf-in: Free-form nerf inpainting with rgb-d priors.
\newblock \emph{arXiv preprint arXiv:2206.04901}, 2022.

\bibitem[Liu et~al.(2021)Liu, Zhang, Zhang, Zhang, Zhu, and Russell]{liu2021editing}
Steven Liu, Xiuming Zhang, Zhoutong Zhang, Richard Zhang, Jun-Yan Zhu, and Bryan Russell.
\newblock Editing conditional radiance fields.
\newblock In \emph{ICCV}, 2021.

\bibitem[Liu et~al.(2023)Liu, Zeng, Ren, Li, Zhang, Yang, Li, Yang, Su, Zhu, et~al.]{liu2023grounding}
Shilong Liu, Zhaoyang Zeng, Tianhe Ren, Feng Li, Hao Zhang, Jie Yang, Chunyuan Li, Jianwei Yang, Hang Su, Jun Zhu, et~al.
\newblock Grounding dino: Marrying dino with grounded pre-training for open-set object detection.
\newblock \emph{arXiv preprint arXiv:2303.05499}, 2023.

\bibitem[Luiten et~al.(2023)Luiten, Kopanas, Leibe, and Ramanan]{luiten2023dynamic}
Jonathon Luiten, Georgios Kopanas, Bastian Leibe, and Deva Ramanan.
\newblock Dynamic 3d gaussians: Tracking by persistent dynamic view synthesis.
\newblock \emph{arXiv preprint arXiv:2308.09713}, 2023.

\bibitem[Michel et~al.(2022)Michel, Bar-On, Liu, Benaim, and Hanocka]{michel2022text2mesh}
Oscar Michel, Roi Bar-On, Richard Liu, Sagie Benaim, and Rana Hanocka.
\newblock Text2mesh: Text-driven neural stylization for meshes.
\newblock In \emph{CVPR}, 2022.

\bibitem[Mildenhall et~al.(2021)Mildenhall, Srinivasan, Tancik, Barron, Ramamoorthi, and Ng]{mildenhall2021nerf}
Ben Mildenhall, Pratul~P Srinivasan, Matthew Tancik, Jonathan~T Barron, Ravi Ramamoorthi, and Ren Ng.
\newblock Nerf: Representing scenes as neural radiance fields for view synthesis.
\newblock \emph{Communications of the ACM}, 2021.

\bibitem[Mirzaei et~al.(2023)Mirzaei, Aumentado-Armstrong, Brubaker, Kelly, Levinshtein, Derpanis, and Gilitschenski]{mirzaei2023watchyoursteps}
Ashkan Mirzaei, Tristan Aumentado-Armstrong, Marcus~A. Brubaker, Jonathan Kelly, Alex Levinshtein, Konstantinos~G. Derpanis, and Igor Gilitschenski.
\newblock Watch your steps: Local image and scene editing by text instructions.
\newblock In \emph{arXiv preprint arXiv:2308.08947}, 2023.

\bibitem[M{\"u}ller et~al.(2022)M{\"u}ller, Evans, Schied, and Keller]{muller2022instant}
Thomas M{\"u}ller, Alex Evans, Christoph Schied, and Alexander Keller.
\newblock Instant neural graphics primitives with a multiresolution hash encoding.
\newblock \emph{ACM Transactions on Graphics (ToG)}, 2022.

\bibitem[Nguyen-Phuoc et~al.(2022)Nguyen-Phuoc, Liu, and Xiao]{nguyen2022snerf}
Thu Nguyen-Phuoc, Feng Liu, and Lei Xiao.
\newblock Snerf: stylized neural implicit representations for 3d scenes.
\newblock \emph{arXiv preprint arXiv:2207.02363}, 2022.

\bibitem[Nichol et~al.(2021)Nichol, Dhariwal, Ramesh, Shyam, Mishkin, McGrew, Sutskever, and Chen]{nichol2021glide}
Alex Nichol, Prafulla Dhariwal, Aditya Ramesh, Pranav Shyam, Pamela Mishkin, Bob McGrew, Ilya Sutskever, and Mark Chen.
\newblock Glide: Towards photorealistic image generation and editing with text-guided diffusion models.
\newblock \emph{arXiv preprint arXiv:2112.10741}, 2021.

\bibitem[Noguchi et~al.(2021)Noguchi, Sun, Lin, and Harada]{noguchi2021neural}
Atsuhiro Noguchi, Xiao Sun, Stephen Lin, and Tatsuya Harada.
\newblock Neural articulated radiance field.
\newblock In \emph{ICCV}, 2021.

\bibitem[Paszke et~al.(2019)Paszke, Gross, Massa, Lerer, Bradbury, Chanan, Killeen, Lin, Gimelshein, Antiga, et~al.]{paszke2019pytorch}
Adam Paszke, Sam Gross, Francisco Massa, Adam Lerer, James Bradbury, Gregory Chanan, Trevor Killeen, Zeming Lin, Natalia Gimelshein, Luca Antiga, et~al.
\newblock Pytorch: An imperative style, high-performance deep learning library.
\newblock \emph{NeurIPS}, 2019.

\bibitem[Radford et~al.(2021)Radford, Kim, Hallacy, Ramesh, Goh, Agarwal, Sastry, Askell, Mishkin, Clark, Krueger, and Sutskever]{Radford2021LearningTV}
Alec Radford, Jong~Wook Kim, Chris Hallacy, A. Ramesh, Gabriel Goh, Sandhini Agarwal, Girish Sastry, Amanda Askell, Pamela Mishkin, Jack Clark, Gretchen Krueger, and Ilya Sutskever.
\newblock Learning transferable visual models from natural language supervision.
\newblock In \emph{ICML}, 2021.

\bibitem[Ramesh et~al.(2022)Ramesh, Dhariwal, Nichol, Chu, and Chen]{ramesh2022hierarchical}
Aditya Ramesh, Prafulla Dhariwal, Alex Nichol, Casey Chu, and Mark Chen.
\newblock Hierarchical text-conditional image generation with clip latents.
\newblock \emph{arXiv preprint arXiv:2204.06125}, 2022.

\bibitem[Rombach et~al.(2022)Rombach, Blattmann, Lorenz, Esser, and Ommer]{rombach2022high}
Robin Rombach, Andreas Blattmann, Dominik Lorenz, Patrick Esser, and Bj{\"o}rn Ommer.
\newblock High-resolution image synthesis with latent diffusion models.
\newblock In \emph{CVPR}, 2022.

\bibitem[Ruiz et~al.(2023)Ruiz, Li, Jampani, Pritch, Rubinstein, and Aberman]{ruiz2023dreambooth}
Nataniel Ruiz, Yuanzhen Li, Varun Jampani, Yael Pritch, Michael Rubinstein, and Kfir Aberman.
\newblock Dreambooth: Fine tuning text-to-image diffusion models for subject-driven generation.
\newblock In \emph{CVPR}, 2023.

\bibitem[Saharia et~al.(2022{\natexlab{a}})Saharia, Chan, Chang, Lee, Ho, Salimans, Fleet, and Norouzi]{saharia2022palette}
Chitwan Saharia, William Chan, Huiwen Chang, Chris Lee, Jonathan Ho, Tim Salimans, David Fleet, and Mohammad Norouzi.
\newblock Palette: Image-to-image diffusion models.
\newblock In \emph{ACM SIGGRAPH 2022 Conference Proceedings}, 2022{\natexlab{a}}.

\bibitem[Saharia et~al.(2022{\natexlab{b}})Saharia, Chan, Saxena, Li, Whang, Denton, Ghasemipour, Gontijo~Lopes, Karagol~Ayan, Salimans, et~al.]{saharia2022photorealistic}
Chitwan Saharia, William Chan, Saurabh Saxena, Lala Li, Jay Whang, Emily~L Denton, Kamyar Ghasemipour, Raphael Gontijo~Lopes, Burcu Karagol~Ayan, Tim Salimans, et~al.
\newblock Photorealistic text-to-image diffusion models with deep language understanding.
\newblock \emph{Advances in Neural Information Processing Systems}, 2022{\natexlab{b}}.

\bibitem[Saharia et~al.(2022{\natexlab{c}})Saharia, Ho, Chan, Salimans, Fleet, and Norouzi]{saharia2022image}
Chitwan Saharia, Jonathan Ho, William Chan, Tim Salimans, David~J Fleet, and Mohammad Norouzi.
\newblock Image super-resolution via iterative refinement.
\newblock \emph{IEEE Transactions on Pattern Analysis and Machine Intelligence}, 2022{\natexlab{c}}.

\bibitem[{Sara Fridovich-Keil and Alex Yu} et~al.(2022){Sara Fridovich-Keil and Alex Yu}, Tancik, Chen, Recht, and Kanazawa]{yu_and_fridovichkeil2021plenoxels}
{Sara Fridovich-Keil and Alex Yu}, Matthew Tancik, Qinhong Chen, Benjamin Recht, and Angjoo Kanazawa.
\newblock Plenoxels: Radiance fields without neural networks.
\newblock In \emph{CVPR}, 2022.

\bibitem[Sohl-Dickstein et~al.(2015)Sohl-Dickstein, Weiss, Maheswaranathan, and Ganguli]{sohl2015deep}
Jascha Sohl-Dickstein, Eric Weiss, Niru Maheswaranathan, and Surya Ganguli.
\newblock Deep unsupervised learning using nonequilibrium thermodynamics.
\newblock In \emph{International conference on machine learning}, 2015.

\bibitem[Song and Ermon(2019)]{song2019generative}
Yang Song and Stefano Ermon.
\newblock Generative modeling by estimating gradients of the data distribution.
\newblock \emph{Advances in neural information processing systems}, 2019.

\bibitem[Sun et~al.(2022)Sun, Sun, and Chen]{sun2022direct}
Cheng Sun, Min Sun, and Hwann-Tzong Chen.
\newblock Direct voxel grid optimization: Super-fast convergence for radiance fields reconstruction.
\newblock In \emph{CVPR}, 2022.

\bibitem[Tschernezki et~al.(2022)Tschernezki, Laina, Larlus, and Vedaldi]{tschernezki2022neural}
Vadim Tschernezki, Iro Laina, Diane Larlus, and Andrea Vedaldi.
\newblock Neural feature fusion fields: 3d distillation of self-supervised 2d image representations.
\newblock In \emph{2022 International Conference on 3D Vision (3DV)}, 2022.

\bibitem[Wang et~al.(2022)Wang, Chai, He, Chen, and Liao]{wang2022clip}
Can Wang, Menglei Chai, Mingming He, Dongdong Chen, and Jing Liao.
\newblock Clip-nerf: Text-and-image driven manipulation of neural radiance fields.
\newblock In \emph{Proceedings of the IEEE/CVF Conference on Computer Vision and Pattern Recognition}, 2022.

\bibitem[Wang et~al.(2023)Wang, Jiang, Chai, He, Chen, and Liao]{nerfart}
Can Wang, Ruixiang Jiang, Menglei Chai, Mingming He, Dongdong Chen, and Jing Liao.
\newblock Nerf-art: Text-driven neural radiance fields stylization.
\newblock \emph{TVCG}, 2023.

\bibitem[Wu et~al.(2023)Wu, Yi, Fang, Xie, Zhang, Wei, Liu, Tian, and Xinggang]{wu20234dgaussians}
Guanjun Wu, Taoran Yi, Jiemin Fang, Lingxi Xie, Xiaopeng Zhang, Wei Wei, Wenyu Liu, Qi Tian, and Wang Xinggang.
\newblock 4d gaussian splatting for real-time dynamic scene rendering.
\newblock \emph{arXiv preprint arXiv:2310.08528}, 2023.

\bibitem[Wu et~al.(2022)Wu, Tan, and Xu]{wu2022palettenerf}
Qiling Wu, Jianchao Tan, and Kun Xu.
\newblock Palettenerf: Palette-based color editing for nerfs.
\newblock \emph{arXiv preprint arXiv:2212.12871}, 2022.

\bibitem[Xu et~al.(2023)Xu, Wang, Cao, Cheng, Shan, and Gao]{xu2023instructp2p}
Jiale Xu, Xintao Wang, Yan-Pei Cao, Weihao Cheng, Ying Shan, and Shenghua Gao.
\newblock Instructp2p: Learning to edit 3d point clouds with text instructions.
\newblock \emph{arXiv preprint arXiv:2306.07154}, 2023.

\bibitem[Xu and Harada(2022)]{xu2022deforming}
Tianhan Xu and Tatsuya Harada.
\newblock Deforming radiance fields with cages.
\newblock In \emph{European Conference on Computer Vision}, 2022.

\bibitem[Yang et~al.(2022)Yang, Bao, Zeng, Bao, Zhang, Cui, and Zhang]{yang2022neumesh}
Bangbang Yang, Chong Bao, Junyi Zeng, Hujun Bao, Yinda Zhang, Zhaopeng Cui, and Guofeng Zhang.
\newblock Neumesh: Learning disentangled neural mesh-based implicit field for geometry and texture editing.
\newblock In \emph{European Conference on Computer Vision}, 2022.

\bibitem[Yang et~al.(2023)Yang, Gao, Zhou, Jiao, Zhang, and Jin]{yang2023deformable3dgs}
Ziyi Yang, Xinyu Gao, Wen Zhou, Shaohui Jiao, Yuqing Zhang, and Xiaogang Jin.
\newblock Deformable 3d gaussians for high-fidelity monocular dynamic scene reconstruction.
\newblock \emph{arXiv preprint arXiv:2309.13101}, 2023.

\bibitem[Yang et~al.(2024)Yang, Yang, Pan, and Zhang]{yang2023gs4d}
Zeyu Yang, Hongye Yang, Zijie Pan, and Li Zhang.
\newblock Real-time photorealistic dynamic scene representation and rendering with 4d gaussian splatting.
\newblock In \emph{ICLR}, 2024.

\bibitem[Yi et~al.(2023)Yi, Fang, Wu, Xie, Zhang, Liu, Tian, and Wang]{GaussianDreamer}
Taoran Yi, Jiemin Fang, Guanjun Wu, Lingxi Xie, Xiaopeng Zhang, Wenyu Liu, Qi Tian, and Xinggang Wang.
\newblock Gaussiandreamer: Fast generation from text to 3d gaussian splatting with point cloud priors.
\newblock \emph{arxiv:2310.08529}, 2023.

\bibitem[Zhang et~al.(2022)Zhang, Kolkin, Bi, Luan, Xu, Shechtman, and Snavely]{zhang2022arf}
Kai Zhang, Nick Kolkin, Sai Bi, Fujun Luan, Zexiang Xu, Eli Shechtman, and Noah Snavely.
\newblock Arf: Artistic radiance fields.
\newblock In \emph{European Conference on Computer Vision}, 2022.

\bibitem[Zhuang et~al.(2023)Zhuang, Wang, Liu, Lin, and Li]{zhuang2023dreameditor}
Jingyu Zhuang, Chen Wang, Lingjie Liu, Liang Lin, and Guanbin Li.
\newblock Dreameditor: Text-driven 3d scene editing with neural fields.
\newblock \emph{arXiv preprint arXiv:2306.13455}, 2023.

\end{thebibliography}
}

\twocolumn[{%
\renewcommand\twocolumn[1][]{#1}%
\maketitle
\vspace{-30pt}
\begin{center}
\centering
\includegraphics[width=1\linewidth]{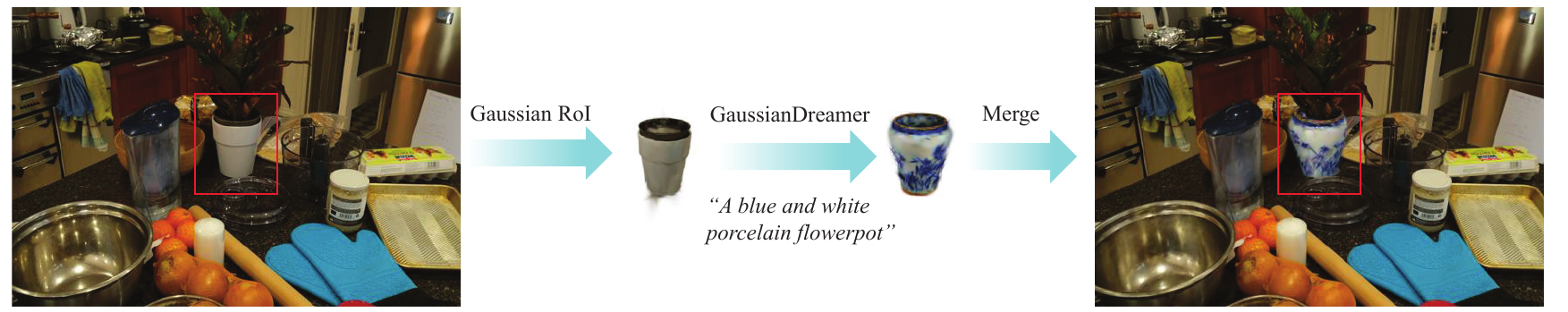}
\vspace{-5pt}
\captionof{figure}{\name demonstrates excellent extension capabilities. It can be seamlessly integrated with the 3D generative model, such as GaussianDreamer~\cite{GaussianDreamer}.
}
\label{fig: GaussianDreamer}
\end{center}%
}]

\appendix

\section{Appendix}

\subsection{Additional Implementation Details}
\name takes a 3D scene reconstructed by 3D Gaussian Splatting~\cite{kerbl20233d} as input. Learning each scene takes 30,000 iterations. Images wider than 512 pixels are resized to 512. Similar to Instruct NeRF2NeRF (IN2N)~\cite{instructnerf2023}, \name also uses Instruct Pix2Pix (IP2P)~\cite{brooks2022instructpix2pix} to edit 2D pictures. The classifier-free diffusion guidance weights are set as follows:
\vspace{5pt}
\begin{itemize}[leftmargin=23pt, itemsep=5pt]
    \item[1)] Fig.~\ref{fig: teaser}: $s_{I}\in[1.4, 1.5], s_{T}\in[7.0, 12.0]$,
    \item[2)] Fig.~\ref{fig: bicycle_and_bear}\:Bicycle: $s_{I}=1.2, s_{T}=12.0$,
    \item[3)] Fig.~\ref{fig: bicycle_and_bear}\:Bear: $s_{I}=1.5, s_{T}=6.5$,
    \item[4)] Fig.~\ref{fig: compare with in2n}: $s_{I}=1.2, s_{T}=8.0$,
    \item[5)] Fig.~\ref{fig: counter}: $s_{I}\in[1.2, 1.5], s_{T}\in[7.5, 12.0]$,
    \item[6)] Fig.~\ref{fig: ablation}: $s_{I}=1.3, s_{T}=12.0$,
\end{itemize}
\vspace{5pt}
where $s_{I}$ is the weight for image guidance and $s_{T}$ is the weight for text guidance.

GaussianEditor implements 3D editing based on the 2D diffusion model. Due to the instability of 2D editing, scenes tend to become blurry as the number of iterations increases. Therefore, we observe the current rendering results during the training process and limit the editing rounds, generally within 200 rounds.

\subsection{Quantitative Evaluation}

\paragraph{Quantitative Evaluation Based on CLIP.}
In Tab.~\ref{tab: quantitative evaluation}, we present the quantitative evaluation results.  The scenes in Fig.~\ref{fig: compare with in2n} are used for this test. We follow the metrics used in Instruct NeRF2NeRF (IN2N)~\cite{instructnerf2023}, including the CLIP~\cite{Radford2021LearningTV} text-image direction similarity and image-image similarity between the original scene and the edited scene. The quantitative results indicate that our method achieves a comparable CLIP text-image direction similarity score with IN2N, while image-image similarity has improved a lot. We would like to analyze the limitations of the used metric as follows.

\begin{table}[htbp]
\centering
\caption{Results of CLIP Text-Image Direction Similarity and Image-Image Similarity between the original scene and edited scene. Test scene is shown in Fig.~\ref{fig: compare with in2n}.}
\begin{tabular}{c|c|c}
\toprule 
  & \begin{tabular}{c} CLIP Text-Image \\ Direction Similarity $\uparrow$ \end{tabular} & \begin{tabular}{c} Image-Image \\Similarity $\uparrow$ \end{tabular} \\
\midrule
IN2N~\cite{instructnerf2023} & \textbf{0.12} & 0.86 \\ 
Ours  &  0.11 & \textbf{0.94} \\
\bottomrule
\end{tabular}
\label{tab: quantitative evaluation}
\end{table}

\begin{figure}[t!]
   \centering
   \includegraphics[width= 0.8\linewidth]{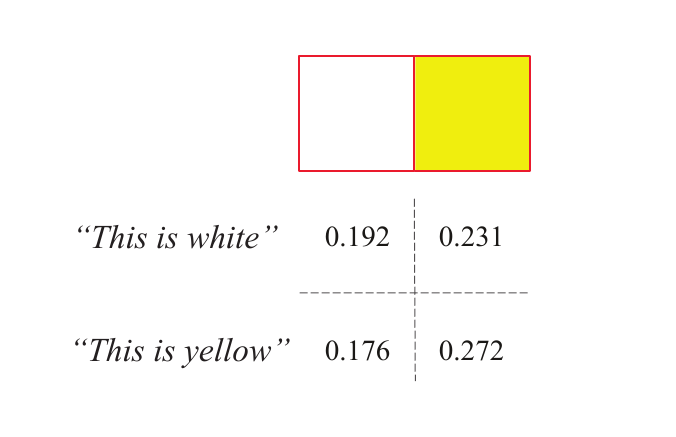}
   \caption{Similarity scores between the text and image features encoded by CLIP~\cite{Radford2021LearningTV}. Pure white images consistently have lower scores\protect\footnotemark.
   }
   \label{fig: limitation_clip}
    \vspace{-12pt}
\end{figure}

\paragraph{Limitation of The CLIP-based Metric.}
Although we provide quantitative analysis based on CLIP. However, we find that the current CLIP-based metrics are not reliable enough. For example, CLIP has problems with color discrimination. As shown in Fig.~\ref{fig: limitation_clip}, we use CLIP to calculate the similarity between solid color images, which are white and yellow respectively, and the text descriptions, \ie ``\texttt{This is white}" or ``\texttt{This is yellow}". The results show that yellow images consistently achieve higher matching scores. This is one of the reasons why our CLIP text-image direction similarity does not show an evident advantage. Therefore, we believe that a more reliable evaluation metric for text-guided editing tasks is one of the important future research directions.

\footnotetext{The red border is to make it easier for readers to see the white image. The actual image input to the CLIP does not have this border.}

\begin{figure}[t!]
   \centering
   \includegraphics[width=\linewidth]{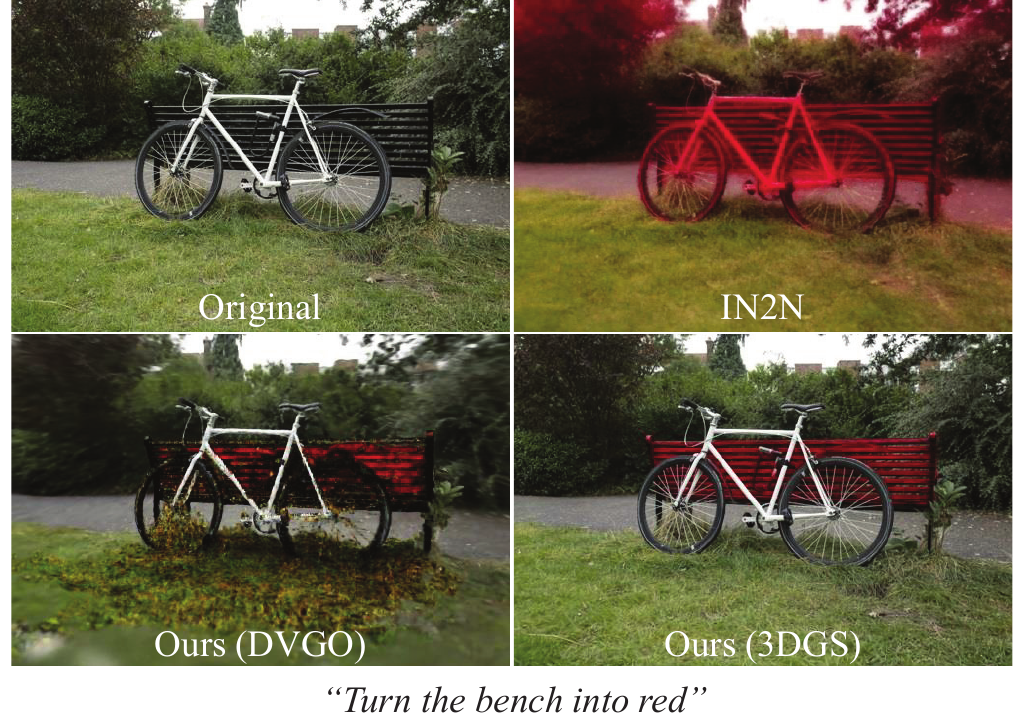}
    \vspace{-15pt}
   \caption{Visualization result of Tab.~\ref{tab: quantitative}.
   }
   \label{fig: mip360}
    \vspace{-10pt}
\end{figure}
\vspace{-5pt}
\paragraph{User Study.} Here are more details of the user study shown in Sec.~\ref{subsec: quantitative}. 4 human editing results in Fig.~\ref{fig: compare with in2n} and 3 bear editing results in Fig.~\ref{fig: bicycle_and_bear} are chosen for the user study, forming 7 questions for the questionnaire. In every question, we showcase the original scene, the text instructions for editing, and the editing results of IN2N~\cite{instructnerf2023} and \name. For equality, the editing results in the question are randomly named using the letter A or B. Users are required to choose the better one. After 21 users submit their questionnaires, 147 votes (21 users $\times$ 7 questions) are collected. \name gets 128 votes for all questions and IN2N gets 19 votes, accounting for 87.07\% and 12.93\%, respectively.

\subsection{Qualitative Evaluation}

\begin{figure}[t!]
   \centering
   \includegraphics[width= \linewidth]{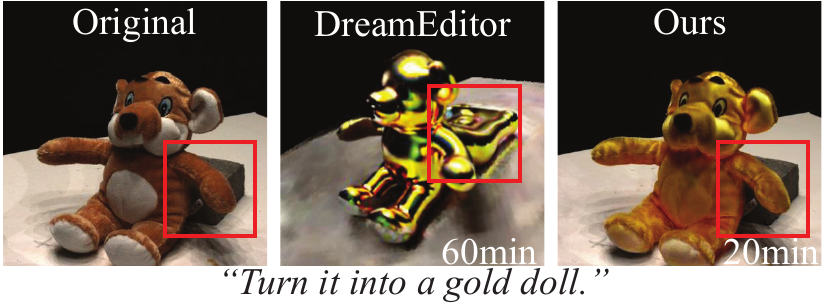}
    \vspace{-20pt}
   \caption{Comparison to DreamEditor on DTU dataset.
   }
   \label{fig: dtu}
\end{figure}

\begin{figure}[t!]
   \centering
   \includegraphics[width= \linewidth]{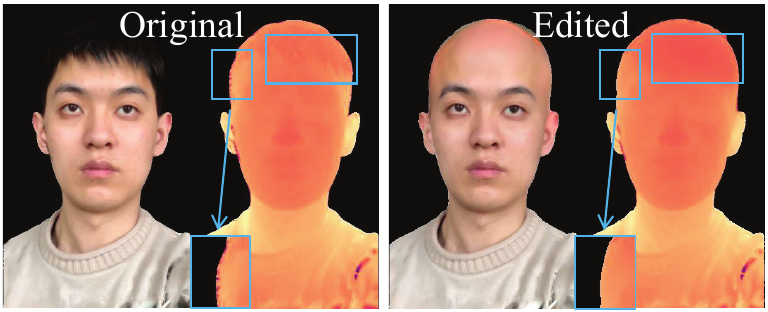}
    \vspace{-15pt}
   \caption{Depth map of hair editing in Fig.\ref{fig: teaser}.
   }
   \label{fig: depth}
    \vspace{-12pt}
\end{figure}

\paragraph{Comparison with IN2N~\cite{instructnerf2023} and Different Backbones.} 
In Fig.~\ref{fig: mip360}, we show the qualitative result of IN2N and \name with different backbones. This scene is also used in Tab.~\ref{tab: quantitative}. IN2N fails in this task and turns the bicycle, bench, and tree all red. Besides, the backbone using DVGO~\cite{sun2022direct} also has difficulty in localizing the bench precisely and produces worse rendering results, while \name grounds the bench precisely and turns it red.
\vspace{-10pt}
\paragraph{Comparison with DreamEditor~\cite{zhuang2023dreameditor}.} In Fig.~\ref{fig: dtu}, we show the qualitative result of DreamEditor and \name. \name delicately edits the doll and retains the hair details, while DreamEditor wipes the hair and changes the back box. Besides, \name gets the wanted editing result using less time.
\vspace{-10pt}
\paragraph{Depth Map of Geometric Editing.} In Fig.~\ref{fig: depth}, we show the depth map of the hair editing result shown in Fig.\ref{fig: teaser}. The depth map indicates that \name possesses a certain level of geometric editing capability. The task of handling drastic geometric editing changes is left for future work. 

\subsection{Extension}
\name demonstrates excellent extension abilities. For instance, it can be seamlessly integrated with the 3D generative model GaussianDreamer~\cite{GaussianDreamer}, resulting in enhanced editing effects. Specifically, as shown in Fig.~\ref{fig: GaussianDreamer}, upon obtaining the Gaussian RoI, the Gaussians within the RoI are saved individually and utilized as the initialization for the 3D-generation model. Simultaneously, the text description of the edited scene is fed into the pipeline of the 3D generation model. Eventually, the edited new object is merged into the original scene to form an edited 3D scene.

\end{document}